\documentclass{article}
\usepackage{amsmath,amssymb,graphicx,mlspconf}
\usepackage{xcolor}
\usepackage{bm}
\usepackage{float}
\usepackage{booktabs}
\usepackage{subcaption, caption}
\usepackage{setspace}
\usepackage{colortbl}
\usepackage[table]{xcolor}

\usepackage{enumitem}

\copyrightnotice{979-8-3503-2411-2/25/\$31.00 {\copyright}2025 IEEE}

\toappear{2025 IEEE International Workshop on Machine Learning for Signal Processing, Aug.\ 31-- Sep.\ 3, 2025, Istanbul, Turkey}

\title{Learning Rate Should Scale Inversely with High-Order Data Moments in High-Dimensional Online Independent Component Analysis}

\name{M. Oğuzhan Gültekin,\; Samet Demir,\; Zafer Doğan\thanks{We acknowledge that this work is supported partially by TÜBİTAK under project 124E063 in ARDEB 1001 program. M.O.G. is supported by the TÜBİTAK project. S.D. is supported by an AI Fellowship provided by KUIS AI Research Center and a PhD Scholarship (BİDEB 2211) from TÜBİTAK. The corresponding author is Zafer Doğan (zdogan@ku.edu.tr).}}
\address{
    MLIP Research Group, KUIS AI Center \& Department of EEE, Koç University\\
    İstanbul, Turkey
}

\begin{document}

\maketitle

\begin{abstract}
We investigate the impact of high-order moments on the learning dynamics of an online Independent Component Analysis (ICA) algorithm under a high-dimensional data model composed of a weighted sum of two non-Gaussian random variables. This model allows precise control of the input moment structure via a weighting parameter. Building on an existing ordinary differential equation (ODE)-based analysis in the high-dimensional limit, we demonstrate that as the high-order moments increase, the algorithm exhibits slower convergence and demands both a lower learning rate and greater initial alignment to achieve informative solutions. Our findings highlight the algorithm’s sensitivity to the statistical structure of the input data, particularly its moment characteristics. Furthermore, the ODE framework reveals a critical learning rate threshold necessary for learning when moments approach their maximum. These insights motivate future directions in moment-aware initialization and adaptive learning rate strategies to counteract the degradation in learning speed caused by high non-Gaussianity, thereby enhancing the robustness and efficiency of ICA in complex, high-dimensional settings.
\end{abstract}

\section{Introduction}\label{sec:intro}
The data encountered in many real-world applications rarely follow idealized Gaussian distributions. Instead, they frequently exhibit non-Gaussian characteristics such as heavy tails, skewness, or higher kurtosis \cite{mahmoudian2019exploring}. These deviations from Gaussianity are more than statistical nuances—they fundamentally alter the behavior of learning algorithms, as meaningful structure must be extracted from higher-order dependencies beyond second-order statistics \cite{lee2000ica, hyvarinen2000independent}. Although non-Gaussianity can enrich the information available for learning tasks, it also introduces significant challenges \cite{ament2024robust, raj2023algorithmic}. Greater departures from Gaussianity can create more complex optimization landscapes, increase sensitivity to noise, and demand more favorable initial conditions to ensure convergence to informative solutions \cite{nguyen2019first, NIPS2016_6ef80bb2, jiao2024emergence}. These effects are particularly pronounced in high-dimensional regimes, where the interaction between algorithm dynamics and data statistics becomes increasingly intricate \cite{advani2020high}. Therefore, understanding the influence of the non-Gaussian structure on algorithmic performance is essential to develop robust and scalable machine learning systems in practice \cite{jin2017optimizationtestinglinearnongaussian}.

Independent Component Analysis (ICA) offers a natural and principled framework for studying the interplay between non-Gaussianity and learning behavior. The goal of ICA is to recover statistically independent latent components from observed linear mixtures, a task that crucially relies on the non-Gaussianity of the source signals \cite{comon1994independent}. Unlike methods that operate solely on second-order correlations, ICA exploits higher-order statistics to enable identifiability and separation \cite{cardoso1999high, zhang2006ica}. This inherent dependence on high-order moments makes ICA particularly well-suited for analyzing how the statistical structure of data affects algorithmic dynamics \cite{rattray2001stochastictrappingsolvablemodel}. When source distributions exhibit strong non-Gaussianity, the learning behavior, sensitivity to initialization, and robustness of ICA algorithms can change dramatically \cite{bhowmik2019enhancement}. As a result, ICA serves as a valuable testbed for investigating how algorithmic behavior is shaped by variations in higher-order statistical features.

Recent theoretical advances have enabled a rigorous study of ICA algorithms in the high-dimensional limit using tools from dynamical systems theory. In particular, it has been shown that the evolution of alignment between estimated and true sources in online ICA can be characterized by deterministic ordinary differential equations (ODEs) \cite{7606821, DBLP:journals/corr/abs-1805-08349}. A notable example is the work of Wang and Lu \cite{10.5555/3295222.3295409}, who derived an ODE-based framework for an online ICA algorithm assuming a single non-Gaussian source component. Their analysis revealed phase transitions in learning behavior, demonstrating that successful recovery depends critically on both the learning rate and the initial alignment. These transitions highlight that even within the same algorithm, changes in the underlying statistical moments can lead to stark differences in learning behavior \cite{auddy2023largedimensionalindependentcomponent}.

In parallel, Ricci et al.\cite{ricci2025feature} developed a quantitative theory of feature learning from non-Gaussian inputs in high-dimensional ICA, focusing on sample complexity and recovery guarantees of the FastICA algorithm. Their work established how the non-Gaussian structure of the sources affects the number of samples needed for reliable extraction. However, their analysis assumes a fixed latent distribution and does not explore how systematic variations in high-order moments impact learning trajectories, convergence thresholds, or the tuning of learning parameters. Thus, a key open question remains: How does the dynamic behavior of online ICA algorithms change when the degree of non-Gaussianity in the data is itself controllable?

Motivated by this question, we extend the classical online ICA setting by introducing a flexible data model composed of a weighted sum of two non-Gaussian random variables. This construction enables fine-grained control over the high-order moments of the input signal via a continuous weighting parameter. Embedded within an ODE-based high-dimensional analysis, our framework allows us to systematically investigate how learning dynamics evolve as a function of moment structure, initialization quality, and learning rate. We find that as the high-order moments increase, the algorithm becomes more fragile: convergence to an informative solution slows, smaller learning rates are required, and a higher initial alignment is necessary to reach informative solutions. These findings reveal a fundamental trade-off between statistical richness and algorithmic stability, emphasizing that increased non-Gaussianity—while beneficial for identifiability—can complicate the learning dynamics of ICA in high-dimensional settings. Our contributions are as follows:
\begin{enumerate}[topsep=0pt,itemsep=-1ex,partopsep=1ex,parsep=1ex,leftmargin=0.5cm]
\item We introduce a high-dimensional ICA data model with controllable high-order moments via a weighted composition of non-Gaussian sources.
\item We extend ODE-based analysis to study how moment structure, learning rate, and initialization affect learning dynamics.
\item We reveal a trade-off between non-Gaussianity and stability: higher moments slow convergence and increase sensitivity to learning parameters.
\end{enumerate}

\textbf{Notations:} Throughout this paper, bold lowercase letters are used, such as $\bm{u}$ and $\bm{x}_k$, to represent \textit{n}-dimensional vectors. The subscript \textit{k} in $\bm{x}_k$ refers to the discrete-time iteration index. The \textit{i}th component of the vectors $\bm{u}$ and $\bm{x}_k$ are denoted by $u_i$ and $x_{k,i}$, respectively.

\section{Problem Definition}
We consider a data model and the algorithmic framework for studying online independent component analysis (ICA) in high dimensions. The data is modeled as a sequence of observations $\boldsymbol{y}_k \in \mathbb{R}^n$, for $k = 1, 2, \cdots$, each generated according to 
\begin{equation}
    \bm{y}_k = \frac{1}{\sqrt{n}} \bm{u} c_k + \bm{a}_k,
    \label{eq:datamodel}
\end{equation}
where $\bm{u} \in \mathbb{R}^n$ is a unique feature vector to be recovered, $c_k \in \mathbb{R}$ is an i.i.d random variable drawn from a non-Gaussian distribution with zero mean and unit variance, and $\bm{a}_k \sim \mathcal{N}(0, \bm{I} - \frac{1}{n}\bm{u}\bm{u}^T)$ models the measurement noise. The normalization $\|\bm{u}\| = \sqrt{n}$ ensures that each component $u_i$ remains an $\mathcal{O}(1)$ quantity as $n \to \infty$. The specific choice of noise covariance guarantees that the overall data covariance is identity, i.e., no information about the feature vector $\bm{u}$ can be inferred from second-order statistics alone.

To recover the feature vector $\bm{u}$ from the stream $\{\bm{y}_k\}_{k \geq 1}$, we adopt a computationally efficient online learning algorithm that updates an estimate $\boldsymbol{x}_k$ iteratively using the following projected stochastic gradient descent rule:
\begin{align}
    \tilde{\bm{x}}_k &= \bm{x}_k + \frac{\tau_k}{\sqrt{n}} f\left( \frac{1}{\sqrt{n}} \bm{y}_k^T \bm{x}_k \right) \bm{y}_k - \frac{\tau_k}{n} \phi(\bm{x}_k), 
    \label{eq:algupdt}
    \\\nonumber
    \bm{x}_{k+1} &= \frac{\sqrt{n}}{\| \tilde{\bm{x}}_k \|} \tilde{\bm{x}}_k,
\end{align}
where $f(\cdot)$ is a twice-differentiable nonlinearity tailored to the statistics of $c_k$, and $\phi(\cdot)$ is an element-wise regularization function that encodes prior information about $\bm{u}$. A normalization step is applied at each iteration to maintain the scale of the estimate. This formulation establishes the basis for analyzing the high-dimensional behavior of the algorithm in subsequent sections.

In \cite{10.5555/3295222.3295409}, the asymptotic behavior of the online ICA algorithm described in \eqref{eq:algupdt} is characterized in the high-dimensional limit. As the ambient dimension $n$ tends to infinity, the joint empirical distribution of the true component vector $\boldsymbol{u}$ and the algorithm's estimate $\boldsymbol{x}_k$ is shown to converge weakly to a deterministic measure. This limiting behavior is governed by a nonlinear partial differential equation (PDE) that evolves over time and captures the full distributional dynamics of the algorithm.

Motivated by these results, the goal of this study is to investigate the effect of high-order moments---defined in terms of the statistical properties of $c_k$---on the learning dynamics of online ICA in high dimensions. To this end, we focus on the cosine similarity between the estimate $\bm{x}_k$ and the true feature vector $\bm{u}$, defined as
\begin{align}
    Q_k^n = \frac{1}{n} \bm{u}^T \bm{x}_k,
\end{align}
where the scaling follows from the normalization of both $\bm{u}$ and $\bm{x}_k$ imposed by the algorithm.

Under suitable regularity conditions, once the scaling limit of the empirical measure is established, the discrete-time sequence can be embedded into continuous time by accelerating the time index by a factor of $n$. This corresponds to analyzing the learning process on a longer time scale. Introducing a rescaled time parameter $t$ via $k = \lfloor t n \rfloor$, the limiting cosine similarity can be defined as
\begin{align}
    Q_t := \lim_{n \to \infty} Q_k^n,
\end{align}
which captures the evolution of the alignment between the estimate and the true feature vector in the high-dimensional regime.

Next, we define $q_t = Q_t^2$, for which a governing ordinary differential equation (ODE) was derived in \cite{10.5555/3295222.3295409}. As an illustrative example, we focus on symmetric non-Gaussian sources where $\mathbb{E}[c_k^3] = 0$. Letting $\mathbb{E}[c_k^4] = m_4$ and $\mathbb{E}[c_k^6] = m_6$, and choosing $f(x) = \pm x^3$ and $\phi(x) = 0$, the governing ODE becomes
\begin{align}
    \frac{dq_t}{dt} & = -2 \tau q_t^2 (1 - q_t)(m_4 - 3) 
    \label{eq:ode}
    \\\nonumber
    & - \tau^2 q_t \left[15 q_t^2 (1 - q_t)(m_4 - 3) + q_t^3 (m_6 - 15) + 15 \right].
\end{align}

This result demonstrates that the dynamics of the online ICA algorithm can be simplified to a deterministic ODE that characterizes the evolution of the alignment between the estimated vector and the true feature vector in the high-dimensional limit.

Building on this prior framework, we generalize the data model to include a weighted sum of two independent non-Gaussian random variables as sources. The relative contributions of these components are modulated by a weighting parameter $\beta$, allowing control over the high-order moments of the non-Gaussian data distribution. Despite this extension, the same large-scale analysis remains applicable, and the algorithm's behavior can still be described by an ODE in the limit as the dimension tends to infinity. This generalized setting forms the basis for the subsequent analysis.

\section{Main Results}
In this section, we present our main findings concerning the influence of a moment-controlled non-Gaussian data model on the learning dynamics of a high-dimensional online ICA algorithm. Our study aims to analyze the behavior of the algorithm using the deterministic ordinary differential equation (ODE) in the high-dimensional scaling limit, which characterizes the macroscopic behavior of the algorithm. This ODE offers a tractable yet expressive description of the algorithm’s dynamics and forms the foundation of our analysis.

Notably, the derived ODE is a nonlinear polynomial equation involving three key interacting parameters: the initial alignment $q_0$, the learning rate $\tau$, and the statistical structure of the source distribution, which is modulated by the weighting parameter $\beta$. While the overall algorithmic framework for online ICA remains unchanged, we introduce a novel data model that enables control over the high-order moments of the latent signal through $\beta$, without modifying the algorithm's structural assumptions.

We begin by formally defining this data model and analyzing how variations in $\beta$ affect the fourth and sixth moments of the latent source signal. We then investigate how these moment changes influence the learning dynamics of the ICA algorithm. Finally, we study how increased high-order moments impact both the choice of learning rate and the threshold behavior of $q_0$, by comparing initialization values above and below the analytically derived unstable fixed point.

\subsection{Data model: Weighted sum of two non-gaussian random variables}
The conventional ICA setup considers a single non-Gaussian source. In contrast, our approach models the latent signal as a weighted sum of two distinct non-Gaussian random variables, $Rademacher$ and $Uniform$ distributions, weighted by a parameter $\beta \in [0,1]$. Specifically, the non-Gaussian data signal $c_k$ in our generative data model is defined as
\begin{equation}
    c_k = \beta a_1 + \sqrt{1 - \beta^2} a_2,
\end{equation}
where $a_1 \sim Rademacher$ and $a_2 \sim \mathcal{U}(-\sqrt{3}, \sqrt{3})$. The $Rademacher$ distribution is a discrete distribution that takes values $+1$ and $-1$ with equal probability.
The uniform distribution $\mathcal{U}(-\sqrt{3}, \sqrt{3})$ is a continuous distribution over the interval $[-\sqrt{3}, \sqrt{3}]$ with constant probability density. This interval is chosen so that the uniform distribution has zero mean and unit variance.
This data model ensures that the weighted combination of non-Gaussian random variables retains zero mean and unit variance for all $\beta \in [0,1]$. The stream of samples $\bm{y}_k \in \mathbb{R}^n, k =1,2,...$ are generated according to the data model defined in \eqref{eq:datamodel}. This modification preserves the theoretical properties of the ICA model while introducing a novel mechanism to control the statistical complexity of the latent signal via $\beta$.
\subsection{Moment analysis and its impact on ICA dynamics}
\begin{figure}[h]
    \centering
    \includegraphics[width=1\linewidth]{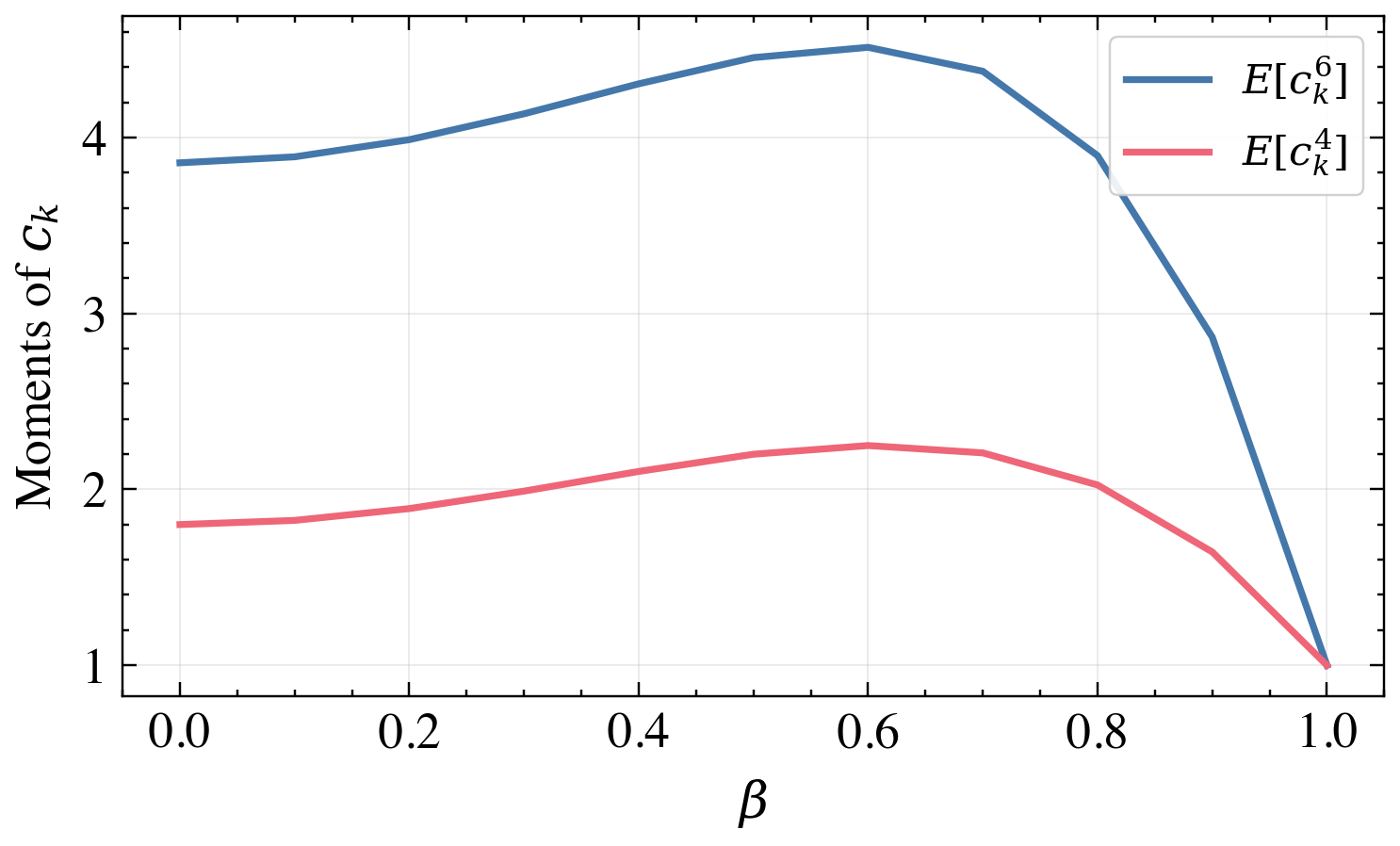}
    \caption{Fourth ($\mathbb{E}c_k^4$) and sixth ($\mathbb{E}c_k^6$) moments of non-Gaussian data $c_k$ with the variation of weigthing parameter $\beta$. Both fourth and sixth moments are maximized at $\beta=0.6$}
    \label{fig:moments}
\end{figure}

The fourth and sixth moments $\mathbb{E}c_k^4=m_4$ and $\mathbb{E}c_k^6=m_6$ of the non-Gaussian data $c_k$, which directly govern the dynamics of the ODE that governs the ICA learning trajectory, are functions of $\beta$. We computed these moments analytically and observed that they are maximized at $\beta = 0.6$ (see Figure \ref{fig:moments}), indicating the mixing regime where the moments can be controlled by $\beta$ for statistical signal content.
\begin{align}
    \mathbb{E}[c_k^4] &= \frac{9}{5} (1 - \beta^2)^2 + 6\beta^2(1 - \beta^2) + \beta^4 \label{eq:fourth_moment} \\
    \mathbb{E}[c_k^6] &= \frac{27}{7}(1 - \beta^2)^3 + 15\beta^2(1 - \beta^2)^2 \label{eq:sixth_moment} \\\nonumber 
    &\quad + 15\beta^4(1 - \beta^2) + \beta^6 
\end{align}
\begin{figure}[h]
    \centering
    \includegraphics[width=1\linewidth]{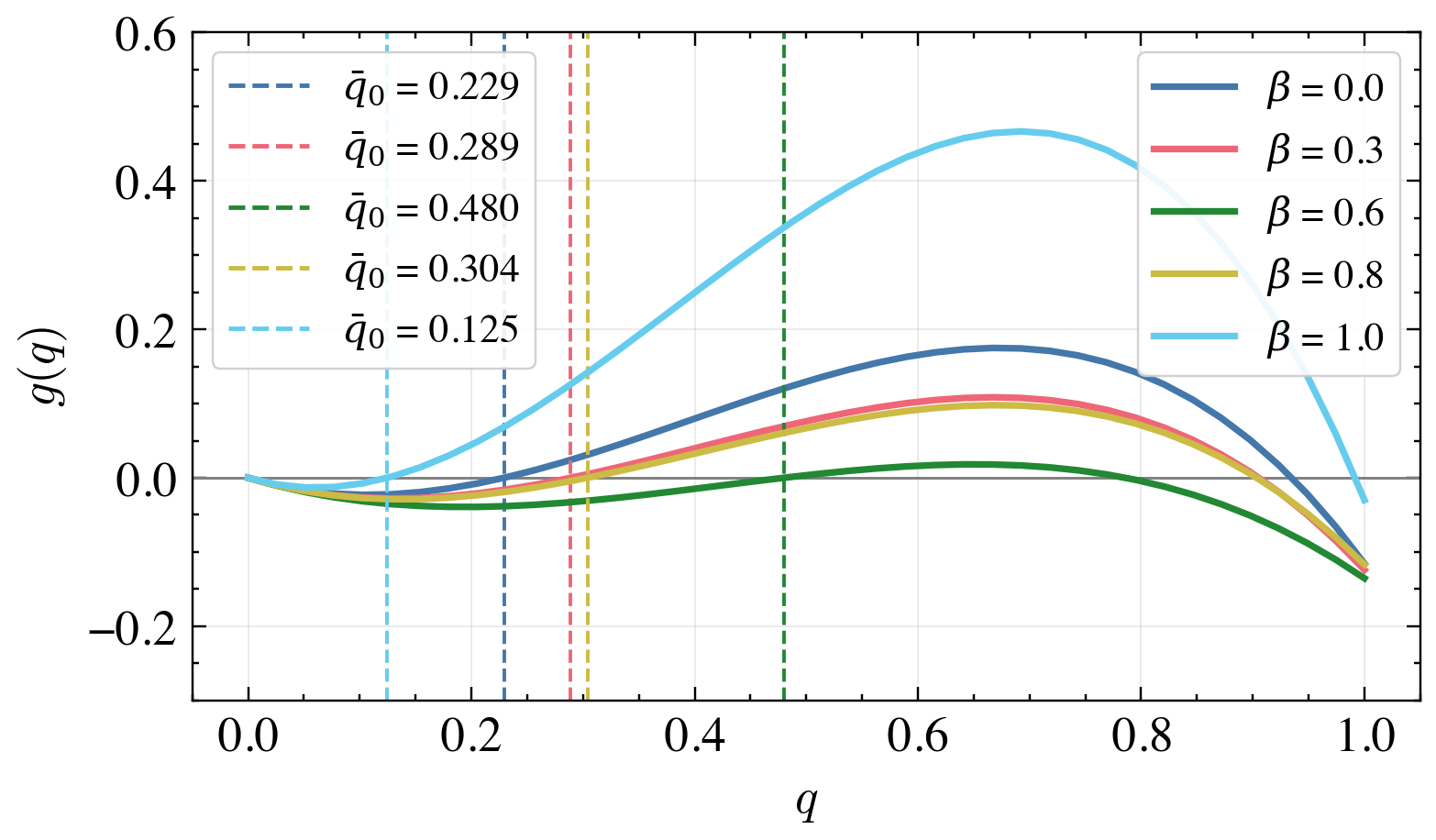}
    \caption{Stability profile of the ODE in \eqref{eq:ode} for $\tau =0.03$, where $g(q)= \frac{1}{\tau}\frac{dq}{dt}$ is plotted for various values of the weighting parameter $\beta \in \{0.0,0.3,0.6,0.8,1.0\}$. The vertical dashed lines indicate the corresponding threshold values $\bar{q}_0$, the smallest positive root of $g(q)=0$, beyond which the informative fixed point becomes locally stable. These thresholds illustrate how the recoverability of the independent component is affected by high-order moments, which is controlled by the signal weighting parameter $\beta$.}
    \label{fig:gq}
\end{figure}
This observation is critical, as the theoretical ODE derived in \eqref{eq:ode} for symmetric non-Gaussian signals with $f(x) = \pm x^3$ and $\phi(x) = 0$ becomes a function of the weighting parameter $\beta$. Since $\beta$ directly modulates the high-order moments of the signal distribution, we emphasize that the effective dynamics and stability of the ICA solution strongly depend on these moments. %
By expressing the fourth and sixth moments of the non-Gaussian data $c_k$ as explicit functions of the weighting parameter $\beta$ in \eqref{eq:fourth_moment} and \eqref{eq:sixth_moment}, we have established a direct link between the statistical structure of the input signals and the stability properties of the learning process. This dependence is particularly important in identifying the regimes under which the algorithm converges to an informative solution successfully.

As we can see from Figure \ref{fig:gq}, the theoretical ODE derived for symmetric data reveals that the learning behavior of the algorithm is strongly influenced by these moments, with maximal instability and sensitivity to initialization occurring at $\beta=0.6$, where the moments are maximized. This moment-induced structure gives rise to phase transitions in the solution landscape, characterized by the emergence of critical initialization threshold and critical learning rate values that separate informative from uninformative fixed points. We investigate this threshold behavior with particular emphasis on how the weighting parameter $\beta$ affects the critical initialization threshold, critical learning rate and the effective speed of convergence.
\begin{figure*}[htp]
    \centering
    \begin{subfigure}{0.49\textwidth}
        \centering
        \includegraphics[width=\linewidth]{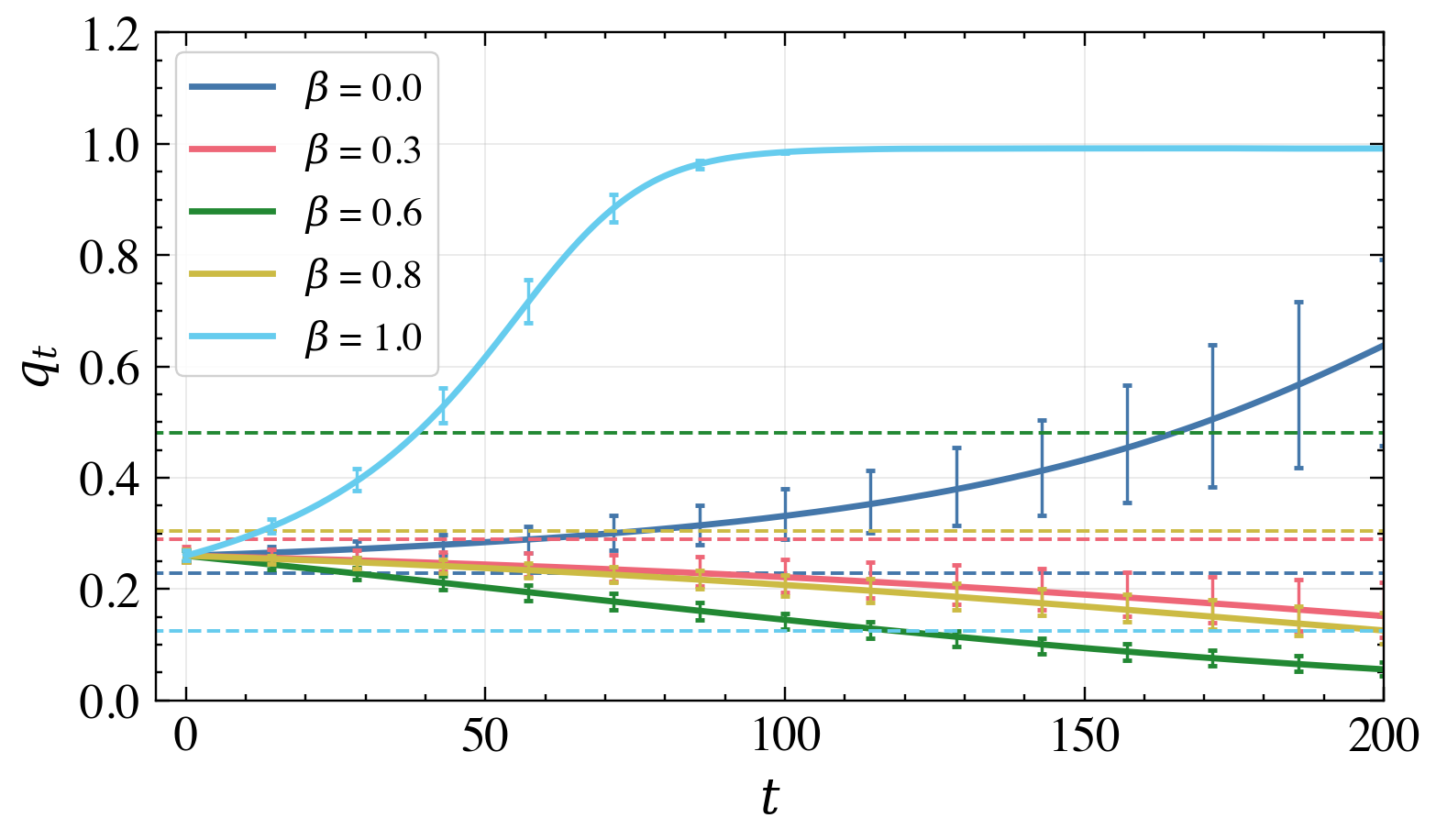}
        \caption{$q_0=0.26$}
        \label{fig:subfig1}
    \end{subfigure}
    \hfill
    \begin{subfigure}{0.49 \textwidth}
        \centering
        \includegraphics[width=\linewidth]{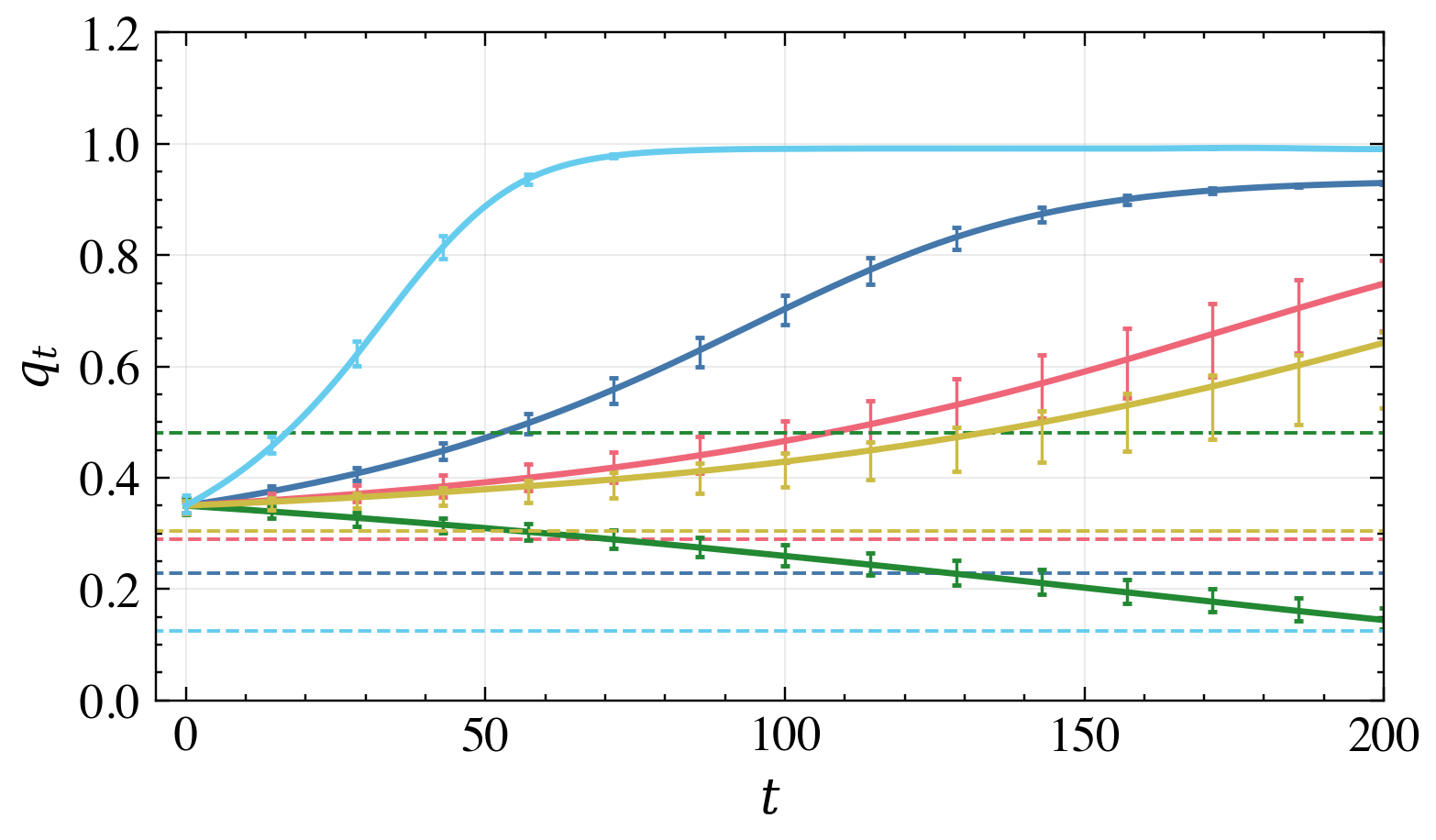}
        \caption{$q_0=0.35$}
        \label{fig:subfig2}
    \end{subfigure}
    \caption{Learning dynamics of the algorithm for different values of $\beta \in \{0.0,0.3,0.6,0.8,1.0\}$, with particular attention to the influence of the initial alignment $q_0$ set either above or below the theoretical threshold. The thresholds are shown by horizontal dashed lines, specific to each $\beta$ (indicated with matched colors). The learning rate is fixed at $\tau = 0.03$. All simulation results are obtained by averaging over 20 independent trials and error bars indicate 1 standard deviation. \textbf{(a)} The initialization is set to $q_0 = 0.26$, which the algorithm goes into the learning regime for the non-Gaussian data corresponding to $\beta \in \{0.0,1.0\}$. In contrast, for all other values of $\beta$, the dynamics fail to converge to an informative solution, resulting in a transition to the uninformative regime. \textbf{(b)} An increased initialization is set to $q_0 = 0.35$, where convergence to an informative solution is observed for the non-Gaussian data with $\beta \in \{0.0,0.3,0.8,1.0\}$, while for the remaining $\beta$ values, the algorithm fails to go into the learning regime.}
    \label{fig:convergence}
\end{figure*}

\subsection{Threshold and learning rate behavior for convergence and learning speed}
To understand the impact of each governing parameter on learning dynamics, we analyze them systematically by varying one while fixing the others. We first define the critical initialization threshold $\bar{q}_0$ as the smallest positive root of the drift function $g(q)$ for a given $\tau$ and $\beta$ values (see Figure~\ref{fig:gq}). This root delineates the boundary beyond which the informative fixed point becomes locally stable. The algorithm converges to the informative solution only when the initial alignment $q_0$ exceeds this threshold, $q_0 > \bar{q}_0$. By computing $\bar{q}_0$ for a fixed $\tau$ and various values of the weighting parameter $\beta$, we define the basin of attraction for successful recovery (see Figure~\ref{fig:convergence}). The threshold shows a non-monotonic trend with respect to $\beta$: it increases up to $\beta = 0.6$ and then decreases. This aligns with our moment analysis, where the fourth and sixth moments peak at $\beta = 0.6$, confirming that higher moments hinder convergence and reduce recoverability in online ICA.

\begin{figure}[h]
    \centering
    \includegraphics[width=1\linewidth]{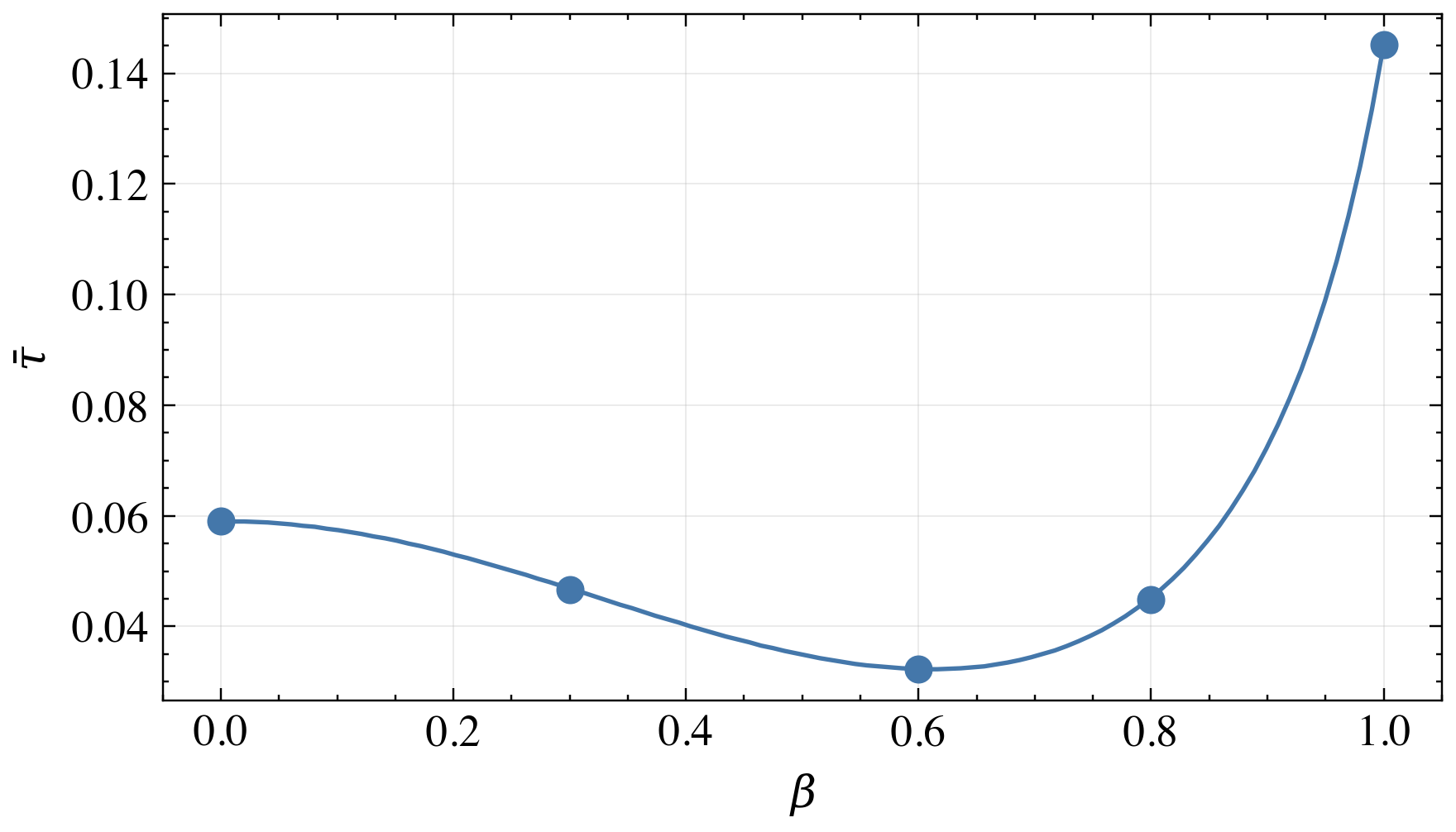}
    \caption{Learning rate thresholds $\bar{\tau}$ for each $\beta$ value such that critical initialization threshold $\bar{q}_0$ for learning exists.}
    \label{fig:cri_tau}
\end{figure}

Next, we examine the influence of the learning rate $\tau$. Numerical solutions of the ODE \eqref{eq:ode} over a grid of $(\tau, \beta)$ pairs reveal a critical threshold $\bar{\tau}$ for each $\beta \in [0,1]$. For $\tau < \bar{\tau}$, convergence to an informative solution is possible when the initial alignment $q_0$ exceeds a threshold; representative values of this threshold are shown in Table~\ref{tab:threshold_q0}. Conversely, when $\tau \geq \bar{\tau}$, the algorithm remains trapped in the uninformative fixed point and does not converge to an informative solution, regardless of initialization. Based on these findings, we observe an inverse relationship between $\bar{\tau}$ and the high-order moments: as the moments increase by varying $\beta$, the algorithm becomes less tolerant to larger $\tau$ values (see Figure~\ref{fig:cri_tau}). This suggests that strong non-Gaussianity, driven by elevated fourth and sixth moments, destabilizes learning under high learning rates.

Importantly, we find that the initialization threshold $\bar{q}_0$ is jointly influenced by both $\beta$ and $\tau$. Higher learning rates reduce the margin of stability for the informative solution and amplify the sensitivity of the algorithm to the initial alignment. Therefore, the learning dynamics are modulated by the interaction between the statistical structure of the source that is controlled with $\beta$ and the algorithm’s learning rate, $\tau$. This highlights a trade-off between learning speed and robustness: while smaller $\tau$ improves stability across a wider range of $\beta$, it may slow down convergence, necessitating careful calibration in practical implementations.

\begin{table}[h]
    \centering
    \caption{Threshold values $\bar{q}_0$ for different ($\tau,\beta$) pairs. Each entry denotes the smallest positive root of the stability function $g(q)=\frac{1}{\tau}\frac{dq}{dt}$, indicating the minimum initial alignment required for convergence to an informative solution. Entries marked as NaN indicate that $g(q)$ has no zero-crossing in the interval $q \in (0,1)$, implying that the ODE remains strictly negative and does not support convergence to an informative solution for the corresponding ($\tau,\beta$) pair. The row highlighted in gray corresponds to $\tau=0.03$ whose stability profiles are visualized in Figure~\ref{fig:gq}.}
    \label{tab:threshold_q0}
    \begin{tabular}{lrrrrr}
    \toprule
    {} & $\beta$ = 0.0 & $\beta$ = 0.3 & $\beta$ = 0.6 & $\beta$ = 0.8 & $\beta$ = 1.0 \\
    \midrule
    $\tau$ = 0.02 & 0.142 & 0.175 & 0.255 & 0.182 & 0.081 \\
    \rowcolor{gray!30}
    $\tau$ = 0.03 & 0.229 & 0.289 & 0.480 & 0.304 & 0.125 \\
    $\tau$ = 0.04 & 0.331 & 0.439 & NaN & 0.470 & 0.171 \\
    $\tau$ = 0.05 & 0.456 & NaN & NaN & NaN & 0.220 \\
    $\tau$ = 0.06 & NaN & NaN & NaN & NaN & 0.270 \\
    \bottomrule
    \end{tabular}
\end{table}

\section{Conclusion}
In this work, we studied the dynamics of an online Independent Component Analysis (ICA) algorithm under a modified data model in which the latent source is expressed as a weighted sum of two non-Gaussian random variables —Rademacher and Uniform—modulated by a weighting parameter $\beta \in [0,1]$. Leveraging a scaling limit characterization of the underlying algorithm-governed by a limiting process characterized as a solution of a nonlinear ODE, we demonstrated that the fourth and sixth moments of the source distribution can be effectively controlled via $\beta$, thereby directly shaping the learning dynamics of the algorithm.

Our findings reveal that increasing high-order moments adversely affects learning behavior: the critical learning rate threshold $\bar{\tau}$ decreases, while the minimum required initialization alignment $\bar{q}_0$ increases. This implies that stronger non-Gaussianity constrains the basin of attraction, making the learning process more sensitive to both the choice of learning rate and the quality of initialization.

These results highlight a fundamental trade-off between statistical richness and algorithmic stability. While non-Gaussianity is necessary for source identifiability in ICA, excessive high-order moments can impede convergence. Future research directions may include the development of adaptive learning rate schedules, robust initialization strategies, or regularization techniques tailored to high-moment regimes, to enhance the stability and efficiency of online ICA in high-dimensional settings.

\bibliographystyle{IEEEbib}
\bibliography{strings,refs}

\end{document}